\documentclass{article}
\usepackage{PRIMEarxiv}
\usepackage{float}
\usepackage[utf8]{inputenc} 
\usepackage[T1]{fontenc}    
\usepackage{hyperref}       
\usepackage{url}            
\usepackage{booktabs}       
\usepackage{amsfonts}       
\usepackage{nicefrac}       
\usepackage{microtype}      
\usepackage{lipsum}
\usepackage{fancyhdr}       
\usepackage{graphicx}       
\graphicspath{{media/}}     
\usepackage{soul,color}
\usepackage{multirow}
\usepackage{amsmath}
\usepackage{amsbsy}
\usepackage{amssymb}

\usepackage{bm}
\title{Exploiting LMM-based knowledge for image classification tasks
\thanks{\textit{\underline{Citation}}: 
\textbf{M. Tzelepi, V. Mezaris, "Exploiting LMM-based knowledge for image classification tasks", Proc. 25th Int. Conf. on Engineering Applications of Neural Networks (EANN/EAAAI 2024), Corfu, Greece, June 2024.}}
\thanks{This preprint has not undergone post-submission improvements or corrections. The Version of Record of this contribution is published in the EANN 2024 Proceedings, CCIS vol. 2141, and is available online at {https://doi.org/10.1007/978-3-031-62495-7\_13}.}
}

\author{
  Maria Tzelepi and Vasileios Mezaris\\ 
  Information Technologies Institute (ITI) \\
  Centre of Research and Technology Hellas (CERTH) \\
  Thessaloniki, Greece\\
  \texttt{\{mtzelepi,bmezaris\}@iti.gr} \\
}

\begin{document}
\maketitle

\begin{abstract}
In this paper we address image classification tasks leveraging knowledge encoded in Large Multimodal Models (LMMs). More specifically, we use the MiniGPT-4 model to extract semantic descriptions for the images, in a multimodal prompting fashion. In the current literature, vision language models such as CLIP, among other approaches, are utilized as feature extractors, using only the image encoder, for solving image classification tasks. In this paper, we propose to additionally use the text encoder to obtain the text embeddings corresponding to the MiniGPT-4-generated semantic descriptions. Thus, we use both the image and text embeddings for solving the image classification task. The experimental evaluation on three datasets validates the improved classification performance achieved by exploiting LMM-based knowledge.
\end{abstract}


%
%
\section{Introduction}
Large Language Models (LLMs) \cite{zhao2023survey,naveed2023comprehensive}, such as GPT-3 \cite{brown2020language} and GPT-4 \cite{achiam2023gpt}, trained on vast amounts of data, have demonstrated exceptional performance in several downstream tasks over the recent few years, placing them squarely at the center of the research activity on Natural Language Processing (NLP) \cite{jiao2023parrot} and computer vision \cite{fu2024drive}. Considering vision recognition downstream tasks, in particular, the emergence of Vision-Language Models (VLMs), such as BLIP-2 \cite{li2023blip} and CLIP \cite{radford2021learning}, allowed for connecting image-based vision models with LLMs, forming Large Multimodal Models (LMMs) (also known as Multimodal Large Language Models) \cite{yin2023survey,zhang2024mm}. For example, MiniGPT-4 \cite{zhu2023minigpt} aligns a frozen visual encoder with a frozen LLM using a single projection layer. Besides, it is noteworthy that CLIP, which apart from achieving remarkable zero-shot performance on various downstream tasks through prompting, is a powerful feature extractor, has been extensively used in the recent literature for various applications \cite{yu2023turning,wu2022wav2clip,ao2023gesturediffuclip}.

In this work, our goal is to address image classification tasks (i.e., tasks of assigning a class label to an image based on its visual content), harnessing the emerging technology of LLMs/LMMs, in order to achieve improved performance in terms of classification accuracy. To achieve this goal, we pursue the direction of utilizing CLIP, proposing to further incorporate knowledge encoded in powerful foundation models such as MiniGPT-4. More specifically, CLIP is commonly utilized as a feature extractor for fitting a linear classifier on the extracted image embeddings and evaluating the performance on various datasets. In this paper, we propose to use MiniGPT-4 for obtaining semantic textual descriptions for each sample of the considered dataset, and then to use the extracted descriptions for feeding them to the textual encoder of CLIP and obtaining the corresponding text embeddings. Thus, instead of using only the image encoder of CLIP to obtain image embeddings, we also use its text encoder to obtain text embeddings that capture LLM-extracted knowledge. As we experimentally show, the extracted LMM knowledge is indeed useful for the downstream image classification task, providing increased classification performance. 

In the recent literature, as discussed in the subsequent section, there have been some efforts for introducing LLM-based knowledge in image and video classification tasks, using the CLIP vision-language model \cite{radford2021learning}. However, to the best of our knowledge, this is the first work that utilizes GPT-4 (MiniGPT-4 in particular) in a multimodal fashion for downstream image classification tasks. More specifically, relevant approaches utilize either the unimodal GPT-3 model for textual prompting (e.g., \cite{yang2023language}) or the multimodal GPT-4 model again for textual prompting (e.g., \cite{maniparambil2023enhancing}). On the contrary, in this work we prompt the MiniGPT-4 model both with text and images. Furthermore, in the literature approaches the LLM is prompted to extract information about the classes of the problem (e.g., \cite{yang2023language}). On the contrary, in this paper, we obtain sample-specific information from the LMM.

The rest of the paper is organized as follows. Section \ref{sec:prior} briefly presents related works. Section \ref{sec:method} presents in detail the proposed methodology for image classification tasks. Subsequently, Section \ref{sec:exp} provides the experimental evaluation of the proposed methodology. Finally, conclusions are drawn in Section \ref{sec:con}.

%
%
\section{Prior Work}\label{sec:prior}
In this section we present previous relevant works, involving CLIP and LLMs / LMMs for classification tasks. First, in \cite{zhou2022learning} the authors address a challenge associated with the deployment of VLMs, such as CLIP, that is, identifying the right prompt. To do so, they propose a method, called Context Optimization (CoOp), for adapting such models for downstream image recognition tasks, exploiting recent advances in prompt learning in NLP. Specifically, they model the context words of a prompt with learnable vectors, keeping the entire pretrained parameters frozen. Next, in \cite{zhou2022conditional} the authors aim to improve the generalization ability of the aforementioned learned context, by proposing a method called Conditional Context Optimization (CoCoOp). More specifically, they extend CoOp by introducing a lightweight neural network so as to produce an input-conditional token for each image, which is combined with the learnable context vectors.

In another recent work \cite{yang2023language}, the authors propose a method called Language Guided Bottlenecks (LaBo) building upon concept bottleneck models \cite{koh2020concept}, aiming to develop interpretable-by-design classifiers. To do so, they combine the GPT-3 language model \cite{brown2020language} and the CLIP vision-language model. More specifically, they prompt GPT-3 to produce candidate concepts describing each of the problem's classes. A submodular optimization is used then to select a subset of discriminating concepts, which are then aligned to images using the CLIP model. A linear layer is applied on the similarity scores of concepts and images in order to learn weight matrix that represents the importance of each concept in the classification task.

Finally, in \cite{maniparambil2023enhancing} the authors utilize the GPT-4 language model \cite{achiam2023gpt} for generating text that is visually descriptive. Then they use this information in order to adapt the CLIP model to downstream tasks. In this way, they can achieve considerable improvements in zero-shot CLIP accuracy, compared to the default CLIP prompt. Furthermore, they propose a self-attention based few-shot adapter, called CLIP-A-self, for exploiting this additional information, providing improved performance over the aforementioned CoCoOp method \cite{zhou2022conditional}. More specifically, CLIP-A-self learns to select and aggregate the most relevant subset of the visually descriptive text in order to produce more generalizable classifiers.

%
%
\section{Proposed Method}\label{sec:method}
In this section we present the proposed pipeline, briefly presenting first the two main components involved in the pipeline, i.e., CLIP and MiniGPT-4.

\subsection{MiniGPT-4}
Following the evolution of LLMs, GPT-4 is the first model to accept both text and image input, producing text output. With the technical details behind GPT-4 remaining undisclosed, MiniGPT-4 was subsequently proposed. MiniGPT-4 aligns a frozen visual encoder with a frozen LLM, utilizing a projection layer. Specifically, MiniGPT-4 uses the Vicuna LLM, while for the visual perception it uses a ViT-G/14 from EVA-CLIP and a Q-Former network. MiniGPT-4 has been utilized in the recent literature for several applications, e.g., \cite{aubakirova2023patfig,zhou2023skingpt,yuan2023artgpt}.

\subsection{CLIP}
Even though VLMs have been emerged since 2015 \cite{vinyals2015show}, CLIP is the first, most popular VLM, trained with natural language supervision at large scale. CLIP comprises of an image encoder and a text encoder, and it is trained with (image, text) pairs for predicting which of the possible (image, text) pairings actually occurred. To do so, it learns a multimodal embedding space by jointly training the image and text encoders to maximize the cosine similarity of the corresponding image and text embeddings of the real pairs in the batch, while minimizing at the same time the cosine similarity of the embeddings of the incorrect pairings. CLIP provides outstanding zero-shot classification performance, using the class labels of the dataset of interest in the text encoder. Another approach to exploit CLIP for classification tasks is to use the CLIP image encoder for extracting the corresponding image representations and use them with a linear classifier (linear probe) achieving exceptional classification performance. Other approaches focus on fine-tuning on a dataset of interest.  

\subsection{Combining image embeddings and MiniGPT-4 descriptions}

\begin{figure}[!h]
  \centering
    \includegraphics[width=\textwidth]{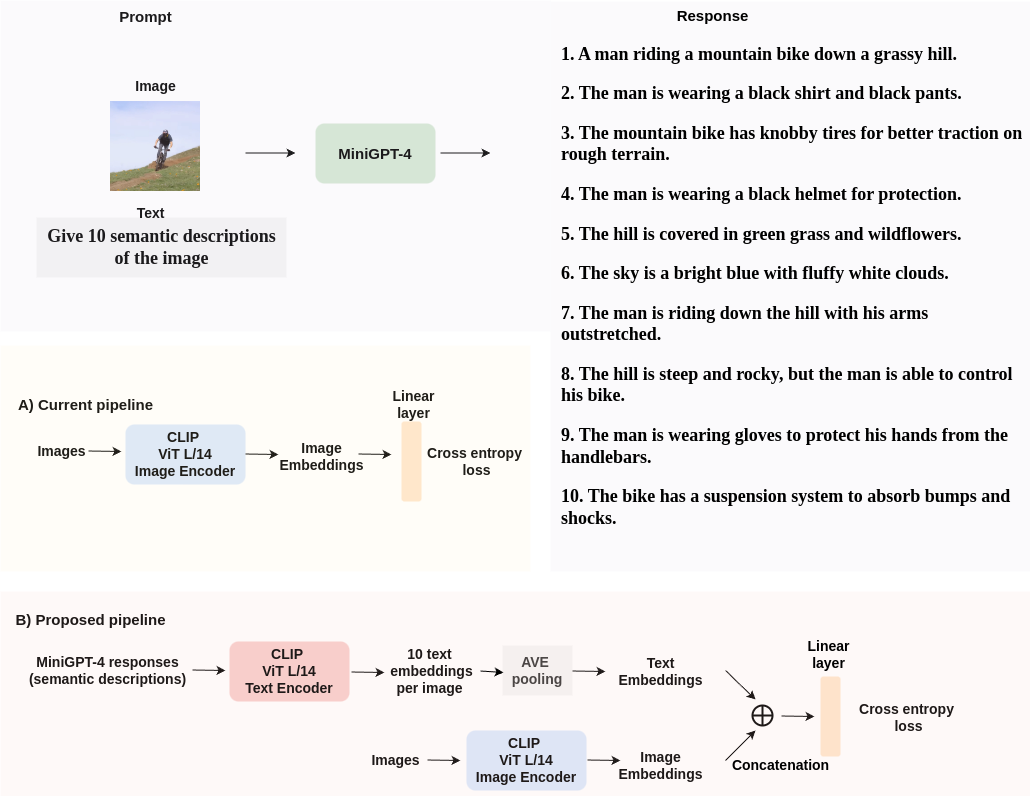}
        \caption{Proposed pipeline for image classification tasks. We first prompt the MiniGPT-4 model for obtaining 10 semantic descriptions of each image of the dataset. While the current pipeline (A) uses the image encoder of CLIP to extract the image embeddings, we propose to exploit the knowledge extracted from the MiniGPT-4 (B). To do so, we also extract the text embeddings of the MiniGPT-4 semantic descriptions using the text encoder of CLIP, followed by the average pooling operation. Finally, the two embeddings for each image are concatenated and propagated to a linear layer for performing the classification task.}
        \label{fig:method}
\end{figure}

In this paper, we aim to exploit knowledge encoded in the MiniGPT-4 model for achieving advanced classification performance in terms of accuracy. To do so, as illustrated in Fig. \ref{fig:method}, we first use the MiniGPT-4 for obtaining semantic descriptions for each image of the dataset. That is, we prompt the model with the image along with the text \textquotedblleft \textit{Give 10 semantic descriptions of the image}\textquotedblright. Subsequently, we introduce the acquired responses to the CLIP's textual encoder (i.e., 10 semantic descriptions for each image of the dataset) and obtain the text embeddings at the final layer of the text encoder. Next, a pooling operation is applied on the aforementioned text embeddings, resulting to a unique embedding for each image. We finally use these embeddings along with the corresponding image embeddings, obtained from the image encoder of CLIP, to feed a single linear layer with a cross entropy loss for performing classification. That is, instead of simply using the image embeddings from the CLIP's image encoder to perform classification, we additionally use the CLIP's text embeddings that capture LMM-based knowledge.

More specifically, let $\mathcal{X} =\{\mathbf{X}_i \in \Re^{h\times w \times c} | i=1, \dots, N\}$ be a set of $N$ images, where $h, w, c$ denote the height, width, and channels of the image respectively. Each image $\mathbf{X}_i$ is linked with a class label $l_i$. Considering a classification task, a common approach is to use the CLIP image encoder $f$ in order to obtain for each image $\mathbf{X}_i$ the corresponding image embedding, $\mathbf{y}_i = f(\mathbf{X}_i) \in \Re^{d}$, where $d$ corresponds to the dimension of the embedding. Then, the image embeddings are used to fit a linear classifier. 

Instead, we use MiniGPT-4 for obtaining semantic descriptions, that is we consider for each image $\mathbf{X}_i$ the MiniGPT-4-generated responses, $d_i^j, j=1,\dots,10$, and then we extract the text embeddings using the CLIP text encoder $g$. That is, $\{\mathbf{t}_i^j=g(d_i^j)  \in \Re^{d}| j=1,\dots,10\}$. Subsequently, the average pooling operation is applied and a unique text embedding $\mathbf{t}^M_i  \in \Re^{d}$ is obtained for each image $\mathbf{X}_i$. For combining the image embeddings with the corresponding text embeddings and realizing the classification task, we generally use concatenation, that is $[\mathbf{y}_i^\intercal,\mathbf{t}^{M\intercal}_i]^\intercal \in \mathbb{R}^{d+d}$. For example, in our experiments we use CLIP with ViT-L-14, were the dimensions of both the image and text embeddings is 768, i.e., $d=768$. The mean embedding, $\mathbf{m}_i \in \Re^{768}$, instead of the concatenated, is also explored. Finally, a learnable linear layer processes the concatenated embeddings to predict the class labels, using the cross entropy loss.

%
%
\section{Experiments}\label{sec:exp}
In this section we present the experiments conducted in order to evaluate the proposed methodology for image classification. First, the utilized datasets are presented followed by the evaluation metrics. Subsequently, the implementation details are presented followed by the experimental results.  

\subsection{Datasets}
We use three datasets to evaluate the effectiveness of the proposed methodology, i.e., UCF-101 \cite{soomro2012ucf101}, Event Recognition in Aerial videos (ERA) \cite{eradataset}, and Biased Action Recognition (BAR) \cite{nam2020learning}. UCF-101 is an action recognition dataset, containing 101 classes. We derive the middle frame from each video and we follow the same protocol as in \cite{soomro2012ucf101} for forming the training and test sets of the 13,320 extracted frames. ERA is an event recognition dataset in unconstrained aerial videos. It consists of 1,473 training images and 1,391 test images, divided into 25 classes. BAR is a real-world image dataset with six action classes which are biased to distinct places, consisting of of 1,941 training images and 654 test images.

\subsection{Evaluation Metrics}
We use test accuracy to evaluate the performance of combining LMM-based features with image features for image classification tasks. Qualitative results of the MiniGPT-4 responses are also provided.

\subsection{Implementation Details}
As previously described, MiniGPT-4 is used as the source of LLM-based knowledge. We use MiniGPT-4 with Vicuna-13B locally. For image and feature extraction we use the ViT-L-14 CLIP version. For the classification task, a single linear layer is used, with output equal to the number of classes of each dataset. Models are trained for 500 epochs, and the learning rate is set to 0.001. Experiments are conducted using the Pytorch framework on an NVIDIA GeForce RTX 3090 with 24 GB of GPU memory.

\subsection{Experimental Results}
\begin{figure*}
\begin{minipage}[b]{0.32\linewidth}
\centering
\includegraphics[width=\textwidth]{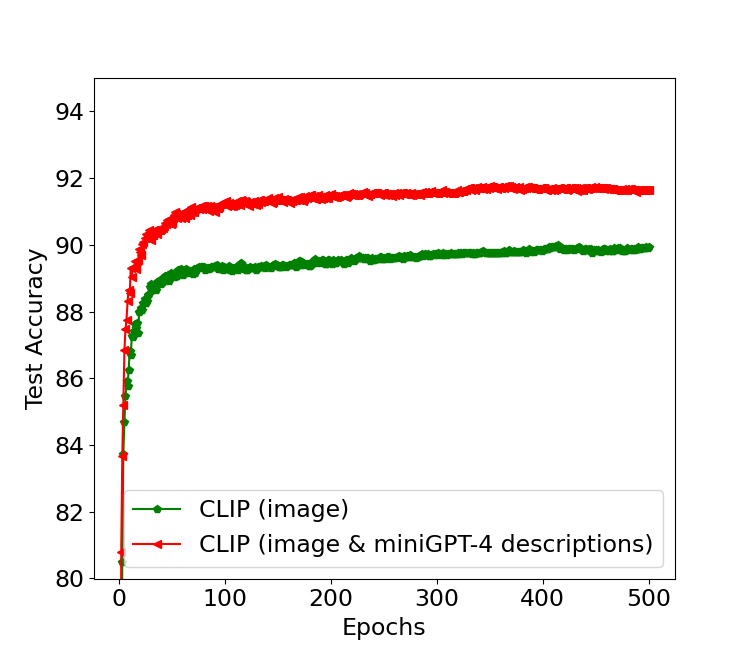}
\caption{UCF dataset: Test accuracy throughout the training epochs.}
\label{fig:ucf}
\end{minipage}
\begin{minipage}[b]{0.32\linewidth}
\centering
\includegraphics[width=\textwidth]{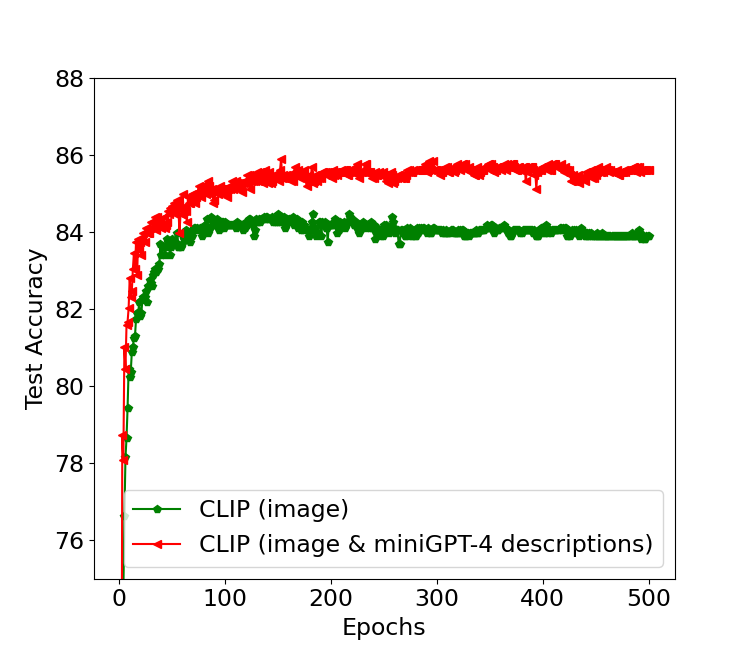}
\caption{ERA dataset: Test accuracy throughout the training epochs.}
\label{fig:era}
\end{minipage}
\begin{minipage}[b]{0.32\linewidth}
\centering
\includegraphics[width=\textwidth]{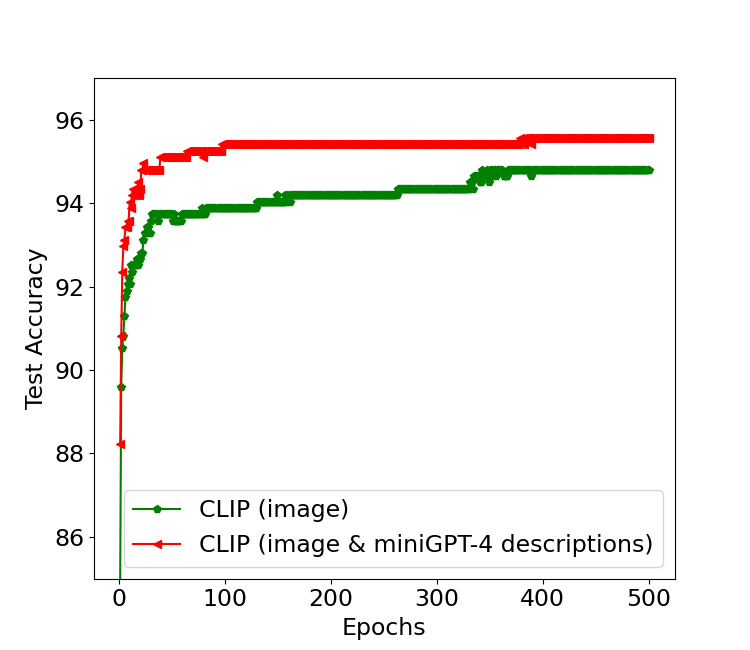}
\caption{BAR dataset: Test accuracy throughout the training epochs.}
\label{fig:bar}
\end{minipage}
\end{figure*}

In this section we present the experimental results of the proposed methodology for addressing image classification tasks. We first provide the evaluation results of the proposed methodology of using both CLIP image embeddings and the CLIP text embeddings that reflect the MiniGPT-4 descriptions for the classification task, against the baseline approach of merely using the CLIP image embeddings, in Table \ref{tab:res}. Best performance is printed in bold. As demonstrated, combining LMM knowledge with image embeddings provides considerably advanced classification performance in all the utilized datasets. Correspondingly, in Fig.s \ref{fig:ucf}-\ref{fig:bar} we provide the test accuracy throughout the training epochs for the two compared approaches on the three utilized datasets, where the steadily advanced performance of the proposed approach is validated. 

\begin{table}
\centering
\caption{Test accuracy.} \label{tab:res}
  \begin{tabular}{|l|l|l|l|}
  \hline
  \bf{Method} & \bf{UCF-101} & \bf{ERA} & \bf{BAR}  \\ 
    \hline
  CLIP (image) \cite{radford2021learning} & 89.981 & 84.472  & 94.801   \\ 
  CLIP (image \& MiniGPT-4 descriptions) & \bf{91.753} & \bf{85.909}  & \bf{95.566} \\ \hline
\end{tabular}
\end{table}

\begin{figure}[H]
\centering
\includegraphics[width=0.75\textwidth]{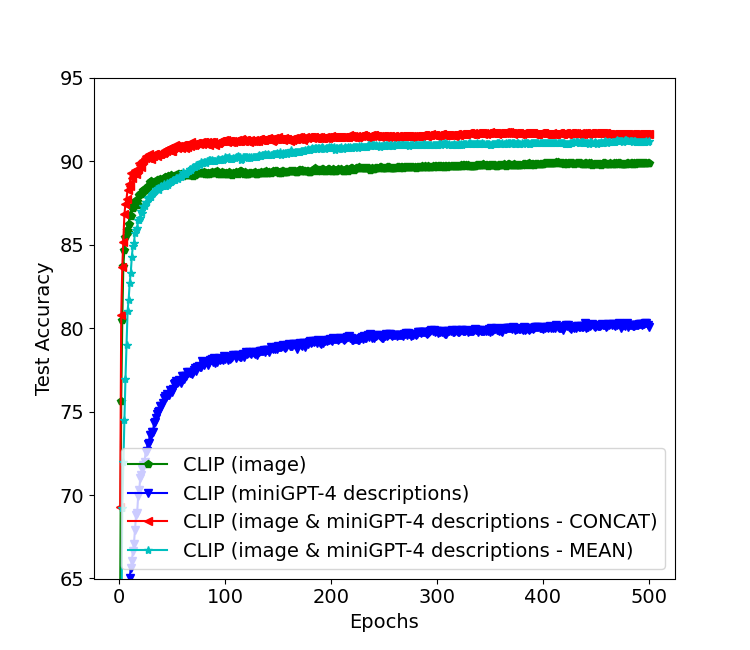}
\caption{Ablation - UCF-101 dataset: Test accuracy throughout the training epochs.}
\label{fig:ucf-ablation}
\end{figure}

Subsequently, an ablation study is performed on the UCF-101 dataset. We present in Table \ref{tab:ablation} the evaluation results. More specifically, we first evaluate the performance of each of the components individually. That is, the classification performance using only the CLIP image embeddings is firstly presented, which serves as the main comparison approach. Evidently, this approach provides exceptional performance. Next, we present the classification performance using only the text embeddings. That is, we only use the MiniGPT-4 knowledge to represent the images of the dataset. This approach performs, as expected, worse, however it achieved remarkable performance. Subsequently, we use both the image and text CLIP embeddings, concatenated. As it is shown, using the proposed methodology we can accomplish significant improvement over the approach of using only the image embeddings. Besides, as it was previously mentioned, we also explore the mean embedding of image and text embeddings instead of the concatenated, which as it is demonstrated also considerably improves the baseline performance, being however slightly worse as compared to the concatenation approach. The aforementioned remarks are validated through Fig. \ref{fig:ucf-ablation}, where the test accuracy throughout the training epochs is provided for all the compared approaches. 

\begin{table}
\centering
\caption{UCF-101: Ablation study.} \label{tab:ablation}
  \begin{tabular}{|l|l|}
  \hline
  \bf{Method} & \bf{Test Accuracy} \\ 
     \hline
  CLIP (image) \cite{radford2021learning} & 89.981 \\ 
  CLIP (MiniGPT-4 descriptions) & 80.333 \\ 
  CLIP (image \& MiniGPT-4 descriptions - CONCAT) & \underline{\bf{91.753}} \\ 
  CLIP (image \& MiniGPT-4 descriptions - MEAN) & \bf{91.277} \\ 
    \hline
\end{tabular}   
\end{table}

In Table \ref{tab:comparisons} we present the comparisons of the proposed methodology against state-of-the-art methods. More specifically, the most relevant and straightforward comparison is against the CLIP model utilizing only the image emebeddings. However, we also include relevant works that include CLIP and LLMs/LMMs. All the models, as described in the Prior Work Section, include training. Finally, even though we do not proceed in a zero-shot approach, we also report the CLIP zero-shot classification performance for completeness. As it is demonstrated the proposed approach achieves the best performance.

\begin{figure}[H]
  \centering
    \includegraphics[width=0.8\textwidth]{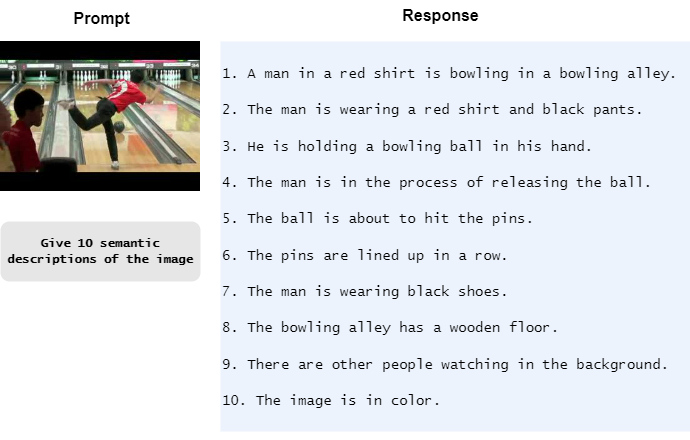}
        \caption{Example of MiniGPT-4 response for a test image of the class \textit{Bowling}.}
        \label{fig:qual1}
\end{figure}

\begin{figure}[H]
  \centering
    \includegraphics[width=0.8\textwidth]{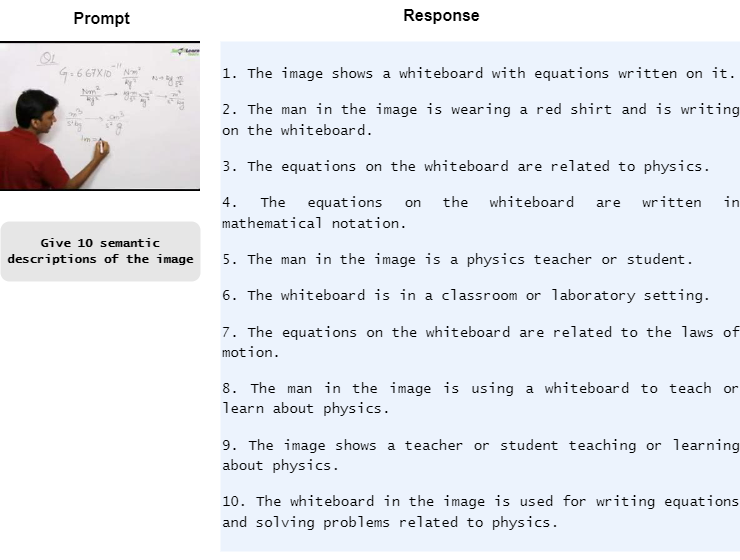}
        \caption{Example of MiniGPT-4 response for a test image of the class \textit{Writing on board}.}
        \label{fig:qual2}
\end{figure}

\begin{figure}[H]
  \centering
    \includegraphics[width=0.8\textwidth]{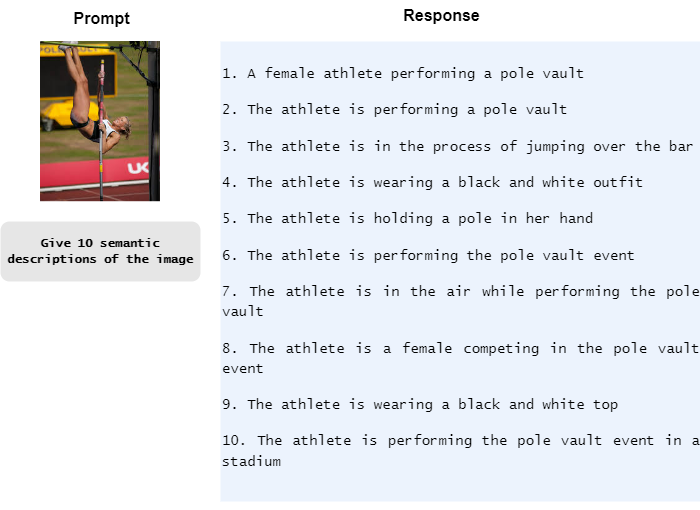}
        \caption{Example of MiniGPT-4 response for a test image of the class \textit{Pole vaulting}.}
        \label{fig:qual3}
\end{figure}

\begin{figure}[H]
  \centering
    \includegraphics[width=0.8\textwidth]{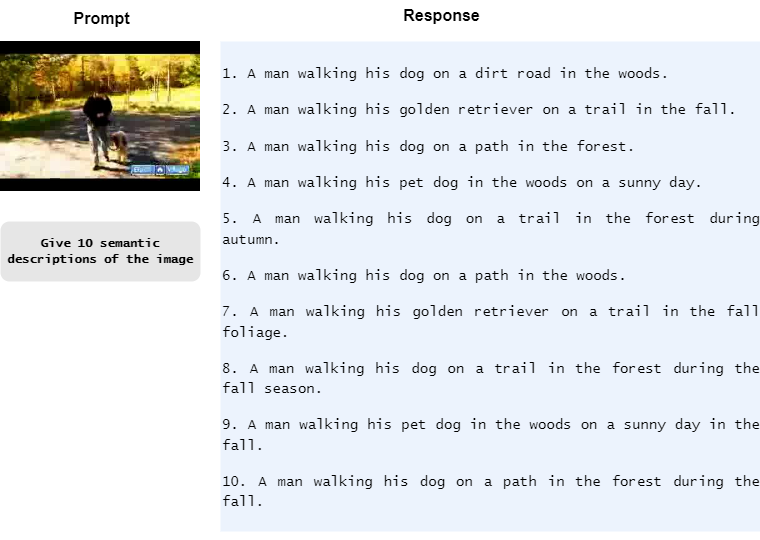}
        \caption{Example of MiniGPT-4 response for a test image of the class \textit{Walking with dog}.}
        \label{fig:qual4}
\end{figure}

Finally, in Figs. \ref{fig:qual1}-\ref{fig:qual4} we provide some qualitative results of the knowledge encoded in the MiniGPT-4 model. More specifically, we present the responses of the model, prompted with the image and the text \textquotedblleft \textit{Give 10 semantic descriptions of the image}\textquotedblright. As it can be observed, accurate and meaningful descriptions are generally obtained.    

\begin{table}
\centering
\caption{UCF-101: Comparisons with CLIP-based approaches.} \label{tab:comparisons}
  \begin{tabular}{|l|l|}
  \hline
  \bf{Method} & \bf{Test Accuracy} \\ 
  \hline
  CLIP-zero shot \cite{radford2021learning}& 76.20 \\
  CLIP-A-self \cite{maniparambil2023enhancing} & 84.1 \\
  CoOp \cite{zhou2022learning} & 84.69 \\ 
  CoCoOp \cite{zhou2022conditional} & 82.33 \\ 
  LaBo \cite{yang2023language} & 90.67 \\ 
  CLIP - Linear Probe \cite{yang2023language} & 90.11  \\ 
  CLIP (image) \cite{radford2021learning} & 89.981 \\ \hline
  CLIP (image \& MiniGPT-4 descriptions) & \bf{91.753} \\
    \hline
  \end{tabular}
\end{table}

%
%
\section{Conclusions}\label{sec:con}
In this paper we deal with image classification tasks exploiting knowledge encoded in LMMs. Specifically, we use the MiniGPT-4 model to extract sample-specific semantic descriptions. Subsequently, we use these description for obtaining the text embeddings from the CLIP's text encoder. Then, we use the aforementioned text embeddings along with the corresponding image embeddings obtained from the CLIP's image encoder for the classification task. As it is experimentally validated, incorporating LMM-based knowledge to the image embeddings achieves considerably improved classification performance.

\section*{Acknowledgements}
This work was supported by the EU Horizon Europe program, under grant agreement 101070109 (TransMIXR).


\end{document}